\pdfoutput=1

\documentclass[11pt]{article}

\usepackage[final]{ACL2023}

\usepackage{times}
\usepackage{latexsym}

\usepackage[T1]{fontenc}
\usepackage[export]{adjustbox}

\usepackage[utf8]{inputenc}

\usepackage{microtype}

\usepackage{inconsolata}

\usepackage{graphicx}
\usepackage{tabularx}
\usepackage{booktabs}
\usepackage{enumitem}
\usepackage{subcaption}
\usepackage{multirow}
\usepackage{hyperref}
\usepackage{listings}

\definecolor{codegreen}{rgb}{0,0.6,0}
\definecolor{codegray}{rgb}{0.5,0.5,0.5}
\definecolor{codepurple}{rgb}{0.58,0,0.82}
\definecolor{backcolour}{rgb}{0.95,0.95,0.92}

\lstdefinestyle{mystyle}{
    backgroundcolor=\color{backcolour},   
    commentstyle=\color{codegreen},
    keywordstyle=\color{magenta},
    numberstyle=\tiny\color{codegray},
    stringstyle=\color{codepurple},
    basicstyle=\ttfamily\footnotesize,
    breakatwhitespace=true,         
    breaklines=true,                 
    captionpos=b,                    
    keepspaces=true,                 
    numbers=left,                    
    numbersep=5pt,                  
    showspaces=false,                
    showstringspaces=false,
    showtabs=false,                  
    tabsize=2
}
\title{Beyond the Battlefield: Framing Analysis of Media Coverage\\ in Conflict Reporting}

\author{
  Avneet Kaur \\ Independent Researcher\And
  Arnav Arora \\
  Independent Researcher}

\begin{document}
\maketitle

\begin{abstract}
Framing used by news media, especially in times of conflict, can have substantial impact on readers' opinion, potentially aggravating the conflict itself. Current studies on the topic of conflict framing have limited insights due to their qualitative nature or only look at surface level generic frames without going deeper. In this work, we identify indicators of war and peace journalism, as outlined by prior work in conflict studies, in a corpus of news articles reporting on the Israel-Palestine war. For our analysis, we use computational approaches, using a combination of frame semantics and large language models to identify both communicative framing and its connection to linguistic framing. Our analysis reveals a higher focus on war based reporting rather than peace based. We also show substantial differences in reporting across the US, UK, and Middle Eastern news outlets in framing who the assailant and victims of the conflict are, surfacing biases within the media. 
\end{abstract}

\section{Introduction}

News media plays an essential role in political discourse, having a substantial impact on shaping public opinion. Trust in news media, however, hinges on unbiased reporting and accurate and complete representation of facts. However, reporting is often biased, either intentionally or unintentionally, due to several factors including individual or collective viewpoints publishers hold, access issues, corporate structures, among others. Analysis and mitigation of such biases is imperative to understand the news media landscape, hold publishers accountable, and establish trust in media institutions. 

To understand these biases in news reporting, journalism and media scholars often employ a technique called framing analysis to uncover patterns in the way a particular topic is reported on. According to the definition given by~\citet{https://doi.org/10.1111/j.1460-2466.1993.tb01304.x}, 
framing is  
\textit{“choosing a few elements of perceived reality and assembling a narrative that highlights connections among them to promote a particular interpretation.”}. Framing analysis, thus, aims to uncover certain salient aspects of an issue that are highlighted over others.

Differences in framing of news can have varied effects on its readers. These can manifest as issue interpretations, cognitive, attitudinal effects and affective effects~\citep{lecheler_news_2019}. These framing effects may in turn shape public opinion and people’s susceptibility to misinformation. In extreme cases, long term exposure to partisan news can reinforce negative stereotyping amongst ethnic groups, polarization of society, inciting hate, intolerance thereby leading to an escalation of conflicts and ideological segregation~\citep{huesmann_foreign_2012, doi:10.1177/10776990221117117, bail-2018-polarisation}.

The dissemination of objective news reporting becomes all the more important during  war and conflict when it has the power to influence political discourse and shape opinion surrounding the war.
Framing a war through a more conflict oriented or a more solution oriented lens can also have substantial impact on public perception. Pioneering work by \citep{Galtung2013HighRL, Galtung1969ViolencePA} established foundational indicators distinguishing war journalism from peace journalism, based on orientations towards peace/conflict, truth, people, and solutions. 

In this work, we utilise Galtung's framework for war vs peace journalism by operationalising each indicator into an NLP-based method and analyse reporting around the ongoing Israel-Palestine conflict~\footnote{\href{https://en.wikipedia.org/wiki/Israeli-Palestinian_conflict}{Israel-Palestine conflict}}. The conflict is one of the most enduring and contentious conflicts, that has continued to cause widespread destruction, loss of innocent lives over several decades. It has gained extensive global media traction over the past couple of years. However, media coverage of the conflict has been under public scrutiny, due to shortcomings related to not providing a nuanced understanding of the conflict~\citep{doi:10.1080/10584609.1993.9962963, Gilboa02092024, neureiter2017sources}, which has continued over several decades.
Several academic studies from communication science, psychology, have also investigated media’s role in crafting the narrative thereby shaping public perception around the conflict~\citep{JoleneFisher,kressel1987biased}.


Most prior work on framing analysis using Galtung's framework is done in communication science through qualitative analysis of news articles, which has limited scalability. Further, computational framing analysis studies do not frame the task in a comparative one, or look into the dynamics around time, geographies and communities~\citep{vallejo-etal-2024-connecting}. 
In this work, we bridge these gaps by operationalising Galtung's framework using Large Language Models (LLM) and classical NLP methods to study editorial patterns in news coverage.  We extract a combination of generic, issue-specific, and semantic frames to identify indicators of War vs. Peace journalism. We identify similarities and differences across media publishers across the UK, USA, and the Middle East w.r.t war reporting by compiling a large-scale dataset of news reports on the conflict from these regions. Through our analysis, we aim to make an effort towards identifying and understand the linguistic structures or framing elements that characterise war and peace journalism in the context of the indicators defined by~\citet{Galtung2013HighRL}. 
To guide our analysis, we aim to answer the following questions:
\begin{itemize}[noitemsep]
    \item How can we operationalise Galtung's framework using NLP methods?
    \item What are the generic, issue-specific, and semantic framing associated with the Israel-Palestine conflict?
    \item Are there differences in how news publishers from different regions frame their coverage of the Israel-Palestine conflict?

\end{itemize}

\section{Background \& Related Work}
\label{sec:background}

In order to understand lexical frame semantics~\citep{fillmore}, i.e. underlying conceptual structures that are associated with words in text, semantic framing, has been studied as a task within NLP. FrameNet~\cite{Baker1998TheBF} is one such framework that has been predominantly used as a resource for semantic analysis, focusing on semantic frames and roles. Several other frameworks and tools such as ~\citet{swayamdipta:17, Kalyanpur2020OpenDomainFS} have been built on top of it and used for automatic semantic frame analysis, aiding in tasks like frame identification and semantic role segmentation. Our approach builds on these approaches, utilising the current state of the art method Frame Semantic Transformer~\citep{chanin2023opensource} for identifying semantic frames.

Media Framing analysis has been done both in the social sciences as well as in NLP. Studies have shown that news media today grapples with inherent biases, manifesting as inadequate or selective reporting of facts, emphasising certain narrative aspects to grasp public attention over the others ~\citep{Morris2007SlantedOP}. Previous work by ~\citet{morstatter_identifying_2018} has analysed news article texts to identify biases in media-reporting. Several efforts have also been made towards uncovering framing biases news reporting war and conflict ~\citep{Baum, Weidmann2016ACL}. Work by ~\citet{Galtung2013HighRL, Galtung1969ViolencePA} on war and peace journalism lays down key indicators and contrasts peace from war journalism based on the difference in focus on violence, propaganda, elite narratives, and victory, characterized by language that demonizes, victimizes, or is emotively charged. In our work, we build on these indicators, outlining an NLP based operationalisation of each indicator.

 While the definition of framing itself is contested, the ~\cite{https://doi.org/10.1111/j.1460-2466.1993.tb01304.x} definition is the most followed one in both the fields. In the social sciences, framing analysis is done to understand a number of topics including, among others, COVID-19~\citep{meyer_framing_2023}, Wall Street~\citep{xu_framing_2013}, Climate Change~\citep{shehata_framing_2012}, and Obesity~\citep{lawrence_framing_2004}. ~\citet{vreese_news_2005} and ~\citet{chong_framing_2007} lay down important frameworks and theory behind detection of frames in general as well as in news media specifically.
Traditionally, journalists and scholars developed and used codebooks manually to identify relevant frames. One such codebook was developed by~\citet{Boydstun2014TrackingTD}. Based on this the Media Frames Corpus was developed by \citep{card-etal-2015-media} and has been widely used as a benchmark to evaluate automated frame classification approaches by the NLP community. Building on those, ~\citet{field_framing_2018} study framing in Russian News while ~\citet{ajjour_modeling_2019} look at frames in argumentation. Several surveys have also been made on computational methods, approaches, formulations and definitions of framing used~\citep{ali_survey_2022, hamborg_automated_2019,Weidmann2016ACL}. There have also been shared tasks to develop and benchmark multilingual framing analysis approaches on a broad range on topics~\citep{piskorski-etal-2023-semeval}, including Israel-Palestine specifically~\citep{zaghouani-etal-2024-fignews} which aims to highlight bias towards either side of the conflict. Work by ~\cite{rosenfeld_fighting_2024} on the conflict uses unigram features to identify user and media bias towards one side. Our work differs from these by utilising a social science framework outlined by~\citep{Galtung2013HighRL} specifically for identifying elements of war vs peace framing and semantic framing, in addition to generic frames, allowing us to do finer grained analysis of nuances specific to conflict reporting and the lexical patterns within it. Our approach moves beyond labels of bias towards one side or the other, outlining the corresponding indicators of war vs peace reporting. 

\section{Data Collection}
For our analysis, we collect a dataset of news articles related to the events surrounding the Israel-Palestine war. We use MediaCloud \citep{roberts2021media} to fetch URLs of articles published b/w October 2023 - February 2024. Since MediaCloud was limited to mid October at the time of data collection, we used the WayBack machine backend to fetch news articles for the month of October. For querying the data source, we use an \textit{"OR"} query with terms such as \textit{"Gaza", "Palestine", "Israel", "Hamas", "Israel Defence Forces"}. We selected a collection of national publishers from the UK, US, and countries from the Middle East. For the Middle East, we chose the countries: \textit{Egypt, Iran, Iraq, Jordan, Lebanon, Qatar, Saudi Arabia, Syria, Turkey, United Arab Emirates, Yemen} which are neighbouring countries or have a political role to play. We explicitly left Israel and Palestinian sources since we wanted to look at how external media is covering the issue.
\subsection{Filtering}
After collecting the dataset, we perform filtering for cleaning the dataset:\newline
\paragraph{Domain} We qualitatively go through the top 20 domains present in the dataset for each region. From these, we manually verified the domains to make sure that it is a reputable news source that is from a country that belongs to the respective region. Based on this analysis, we identified the top domains that are present in the dataset and are reputable. The list of selected domains can be found in Table~\ref{tab:filtered_domains}.

\begin{table}[]
    \centering
    \small
    \begin{tabular}{|p{0.7cm}|p{6cm}|}
    \toprule
    Region & Domains \\
    \midrule
    UK & \texttt{dailymail.co.uk, independent.co.uk, theguardian.com, bbc.co.uk, huffingtonpost.co.uk, telegraph.co.uk}\\
    US & \texttt{nytimes.com, cbsnews.com, foxnews.com, nypost.com, npr.org, breitbart.com}\\
    ME & \texttt{almanar.com.lb, mehrnews.com, egyptindependent.com, cumhuriyet.com.tr, arabnews.com, dohanews.co, sana.sy}\\
    \bottomrule
    \end{tabular}
    \caption{Filtered Domains}
    \label{tab:filtered_domains}
\end{table}

\paragraph{Topic} Even after only downloading the articles based on Israel and Palestine related keywords, we still had some articles about other topics in the dataset. To filter those out, we performed topic modelling on the titles using BERTopic~\citep{grootendorst2022bertopic} with the default parameters and removed the data that was on other unrelated topics, by going though the list of topics qualitatively. For instance, there were many articles on the Israel bombing on Lebanon, which is related but not directly about the conflict. 

\paragraph{Length} Finally, after performing the two steps of filtering, we calculate length of the articles and remove the lowest 1\% and the top 5\% to remove the very short and very long articles from the dataset. We also remove few articles published before 1 October 2023 that were in the dataset despite the range given to MediaCloud.

After the filtering process, we had total of about 22k articles (9.5K US, 8K UK, 4K Middle East). We show the number of articles from each publisher in~\autoref{fig:freq_sources} and distribution of length of articles from each region in \autoref{fig:data-stat-len} in the Appendix.

\begin{table*}[h]
\centering
\small
\begin{tabularx}{\textwidth}{p{2cm}p{4.5cm}p{4.5cm}p{3.5cm}}
\toprule
\textbf{Feature} & \textbf{War Journalism} & \textbf{Peace Journalism} & \textbf{Operationalisation} \\
\midrule
Focus/Orientation & \textbf{Conflict} - use of aggressive, emotive language: "attacks", "genocide",  "victory", "defeat" & \textbf{Solution} - focus towards finding solutions, peace talks, cease fire, use of less aggressive language "dialogue", "peace", "reconciliation" & LLM Based Adversarial Frame and language detection (Attribution of Blame, Demonising Language) \\
\midrule
Effects of War & focus on visible effects: deaths, casualties, destruction, economic damage, use of dramatic representations, focusing on manner of death, "attacks", "killings" "devastation", "tragedy" & focus on invisible effects: suffering, trauma, cultural violence, social disruption, use of subtle, depth reporting, focus on how people are affected: "mental trauma", "suffering", "healing"  &  Semantic Framing + LLM Based extraction of visible/invisible effects of War \\
\midrule
People vs Elite oriented & focus on voice of elites 
  & focus on people groups, voice of common man, women, children & LLM based excerpt extraction of people vs elite mentions\\
  \midrule
Approach & Reactive: focus on responding to violence, Reactive language: "retaliation", "response" & Proactive: focus on preventing violence, Proactive language: "initiatives", "preventive measures" & LLM based detection of solution/ peace orientation\\
\midrule
Time  & immediate events & causes and consequences, long-term & Temporal analysis of events surrounding the conflict\\
\midrule
Dichotomy & Partisan - Biased, partial terminology: "us versus them", "enemy" & Balanced view, Diverse perspectives - Neutral, unbiased terminology: "parties", "stakeholders" & LLM based detection of partisan language\\

\bottomrule
\end{tabularx}
\caption{Conceptualisation and Operationalisation framework for identifying War and Peace Journalism Elements based on work by Galtungs et al.}
\label{tab:journalism_comparison}
\end{table*}

\section{Methodology}
For understanding media framing in a comparative way across regions within our dataset, we conduct analysis at several levels. We conduct high level analysis by extracting generic frames and more fine-grained analysis of communication within the context of war reporting through issue-frames as conceptualised in Galtung's framework. We outline these below:
\subsection{Generic Frames}

For a broad overview of framing, we conduct generic frame analysis of the text in the articles. We use the generic frames listed by the Media Frames Corpus (MFC)~\citep{card-etal-2015-media}, namely: \textit{Economic, Capacity and resources, Morality, Fairness and equality, Legality, constitutionality and jurisprudence, Policy prescription and evaluation, Crime and punishment, Security and defense, Health and safety, Quality of life, Cultural identity, Public opinion, Political, External regulation and reputation, Other}. These provide the framing of an article irrespective of the topic. Using the approach laid out in   \citet{arora2025multimodalframinganalysisnews}, we used an LLM for extracting the framing of the article. Given the text of an article and the above listed frames with their descriptions, we prompt an instruction-tuned Mistral-7b model to generate a list of frames present in the article, along with a reasoning for the classification. 

The model's performance was evaluated on the MFC dataset, calculating overlap in the set of predicted labels and gold labels per article across the 32k articles. The model produced non-zero overlap in 95.7\% of the articles, with an macro-averaged precision and recall over all labels of 0.39 and 0.58 respectively. Considering the challenging and subjective nature of the task of framing analysis and multi-label outputs, the model is found to have competitive performance. We additionally outline metrics per frame label in \autoref{tab:text-frame-perf-label}.

\subsection{Issue-specific frames}

Issue-specific frames are commonly used to interpret specific aspects of a conflict or event. These frames are tailored to the topic and their analysis helps shed light on how audiences perceive and understand the issue at hand. For instance, climate change can be framed as a scientific issue, a political issue, a moral issue, a health issue etc. with frames such as Global Doom, Local Tragedies, Sustainable future etc. For war reporting, Galtung's framework for war and peace journalism has been used study media reporting during times of war, as outlined in Section~\ref{sec:background}. Following that, we operationalise Galtung’s framework to identify indicators of war and peace journalism through NLP methods. In order to capture these indicators in the articles,  we use a combination of semantic framing approaches and LLM identification of communicative framing. 
We list the key indicators that differ between war based vs solution based journalism in \autoref{tab:journalism_comparison}. 

A key pattern in war based journalism is the use demonizing, dehumanizing and victimizing language, which at an extreme level, can be categorized as hate speech or genocidal language. Depending on the leaning of the publisher or the individual journalist, this can advertently or inadvertently lead to adversarial language towards a certain group. We identify the use of such language via an LLM based approach.

Another key aspect is the focus on visible vs invisible effects of war. Media coverage on visible effects of war tends to focus more on visible destruction caused to people, buildings, and its subsequent economic impact. However, they seldom tend to speak about it from a human interest frame. Or, more specifically, the invisible effects of war, such as the  emotional turmoil on people and suffering caused, mental trauma, coping with the aftermath of the war, to name a few. 

We identify these using a combination of semantic framing and an LLM based approach in order to highlight dominant semantic frames as well as the key targets of such language. By identifying the targets, we contrast the differences in reporting across the regions. The list of semantic frames we selected associated to this indicator are outlined in \ref{tab:frames}.
Further, we also capture whether the news article is more people oriented, in terms of reporting, or does it only talk about the interaction of major geo-political entities, leaders etc, highlighting an elite orientation. In addition to that, we capture the presence of diversified perspectives/ voices on the war vs only a single narrative being pushed throughout. 
Finally, we do a temporal analysis of these elements of war, to get an overview of how timely these topics come up in media coverage across various media houses. 

\paragraph{LLM based extraction}
For designing the LLM prompt, as a first step, we iterated over and experimented with an inductive approach, where the LLM was prompted to identify indicators of war based journalism as per Galtung's framework, without elaborate description, or explicitly stating them in the prompt. The resulting output contained indicators such as use of adversarial framing, focus on conflict vs solution oriented journalism. We used some of the descriptive indicators that we got as output to restructure our final prompt in an attempt to make it more explicit and comprehensive. The final prompt is presented in the Appendix. 

For identification of a particular indicator, we prompt the LLM to output the corresponding excerpt from the article, the target of the indicator (where applicable), and the reasoning for the excerpt being an instance of the indicator. By prompting the LLM to provide the corresponding excerpt and a justification grounds the output and reduces hallucinations. We conducted qualitative analysis of the LLM outputs and found them to be faithful to the article.

We used the restructured prompt as input to the LLM and ran it over a sample of 9k articles from our dataset, equally distributed across regions, due to compute limitations. We used Command-R \footnote{\href{https://docs.cohere.com/v2/docs/command-r}{Command-R}} from the Cohere API due to its strong performance on the task during our initial testing.

\subsection{Semantic Frames}
In an attempt to understand semantics of the indicators of war and peace journalism, we apply Frame Semantic parsing over the entire dataset. We do this by using an open source implementation of Frame-Semantic-Transformer model~\citep{chanin2023opensource}, which is a seq-2-seq T5 model trained on FrameNet~\citep{baker-etal-1998-berkeley-framenet}, to all the filtered article text across all regions. We use the model to identify the dominant semantic frames and corresponding frame elements present in a sentence.  We manually go through all the higher level semantic frames in FrameNet as listed on their webpage, to compile a list of relevant frames, or frames of interest, associated with visible and invisible effects of war, people oriented vs elite oriented, time related frames, and those depicting any kind of dichotomisation, and others related to linguistic features associated with Galtung's indicators as mentioned in Table \ref{tab:journalism_comparison}. 



\section{Analysis}

\subsection{Generic framing analysis}
\begin{figure}[h]
    \centering
    \includegraphics[width=\linewidth]{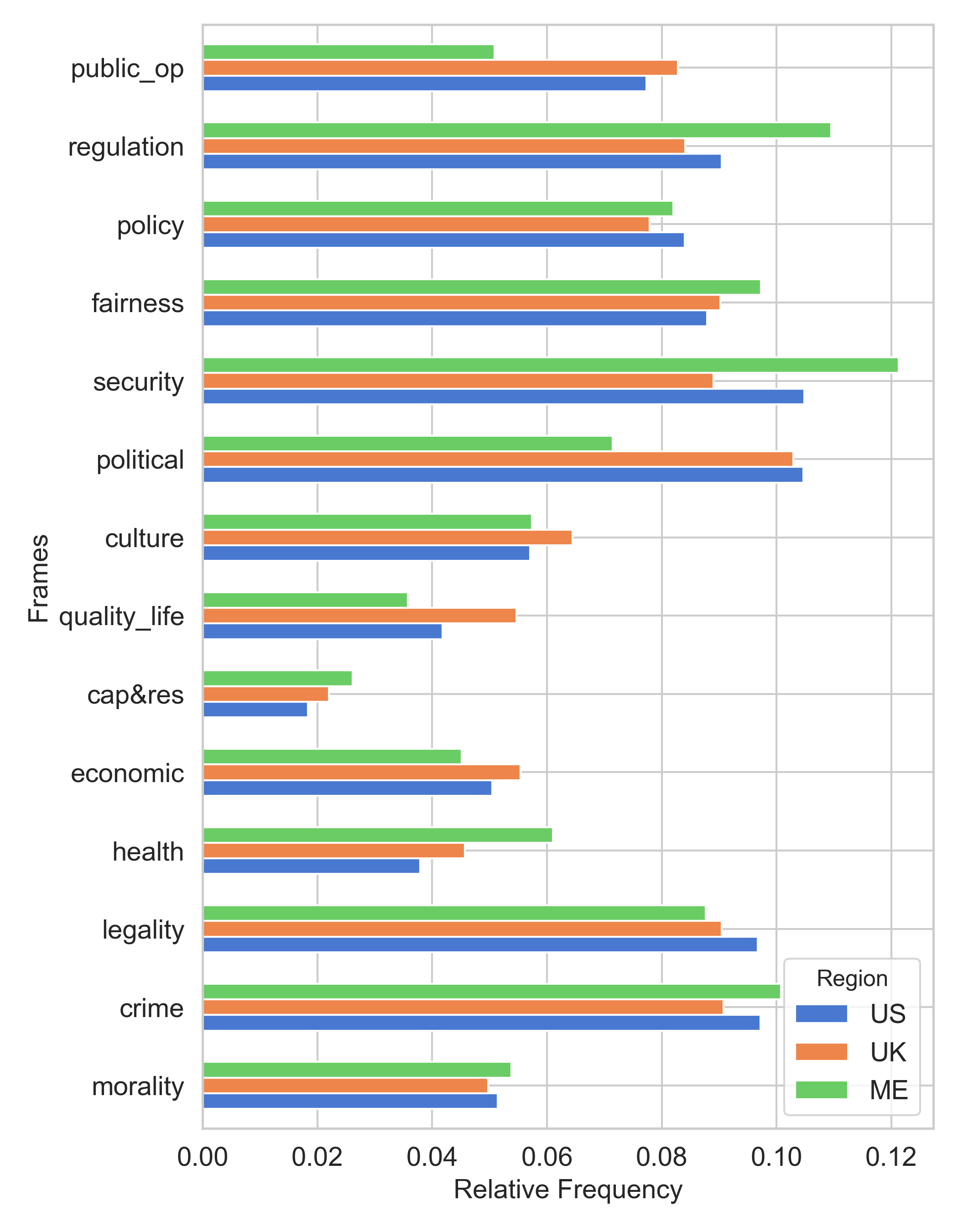}
    \caption{Generic frames per region}
    \label{fig:generic_frame_region}
\end{figure}

We plot the generic framing used by publishers from the different regions in Figure~\ref{fig:generic_frame_region}. Overall, across all the regions, the issue as framed as one of crime, legality, security, and regulation. Publishers in the US and UK are very similar in terms of framing, with the exception of the UK having a higher focus on quality of life and the US on security. The are some key differences between them and the Middle East. Publishers in the Middle East have a much higher focus on health, regulation, and security whereas the US, and UK have much higher focus on framing the issue as one of public opinion and politics.

\subsection{Issue specific Framing Analysis - War and Peace Journalism}

\begin{figure}
    \centering
    \includegraphics[width=\linewidth]{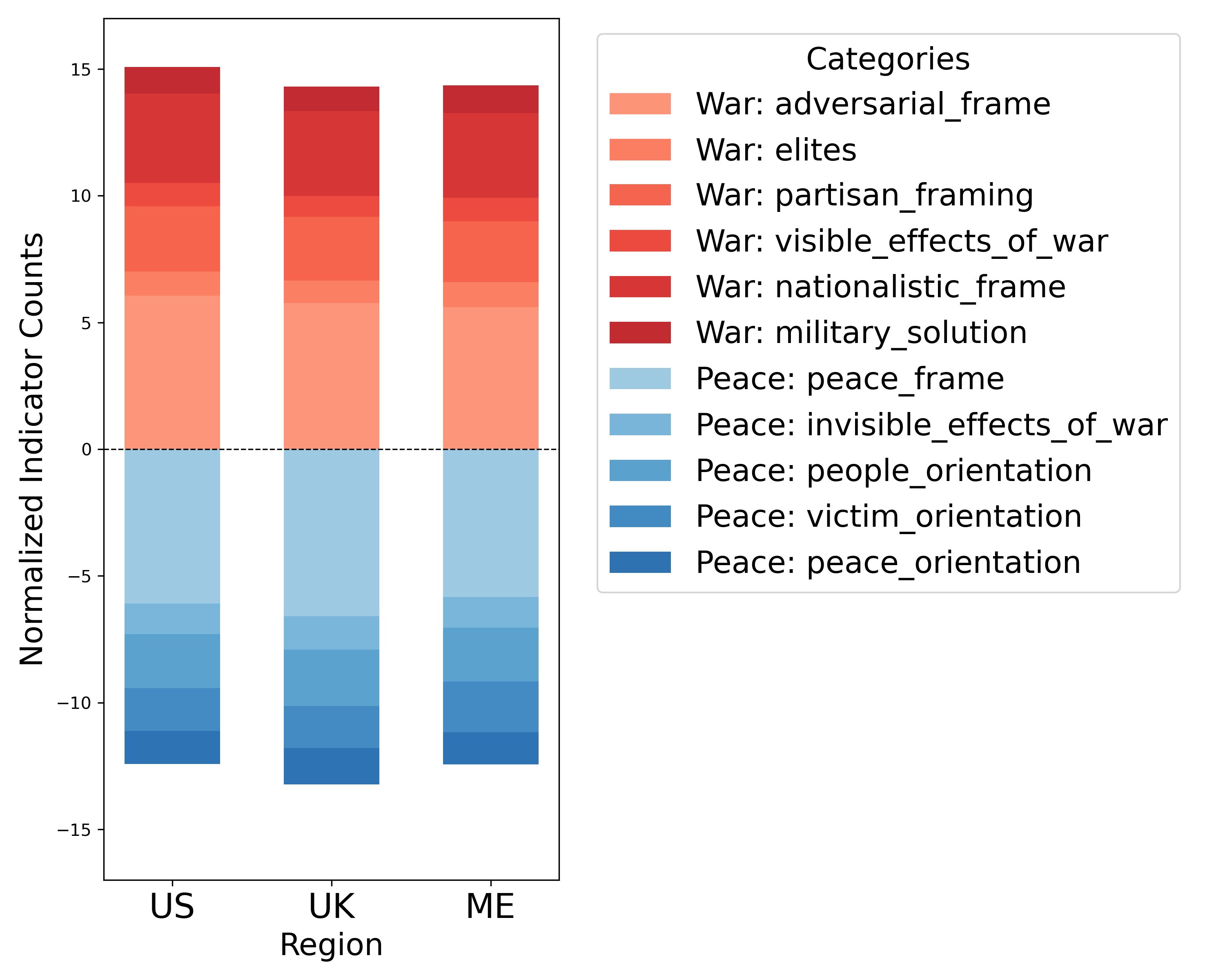}
    \caption{Normalised frequency of war vs peace indicators per region}
    \label{fig:war-peace-freq}
\end{figure}

Going beyond generic frames, we look at different indicators pointing us to war and peace oriented journalism. We plot the frequency normalised by the length of the articles for war and peace indicators in \autoref{fig:war-peace-freq}. We can see that overall, across all three regions, war indicators are found more frequently than peace indicators, demonstrating the higher focus on war reporting. Nationalistic and Adversarial framing are quite common across the dataset whereas there is much lower focus on Invisible effects of war.

\begin{figure}[h]
   \includegraphics[width=0.9\linewidth]{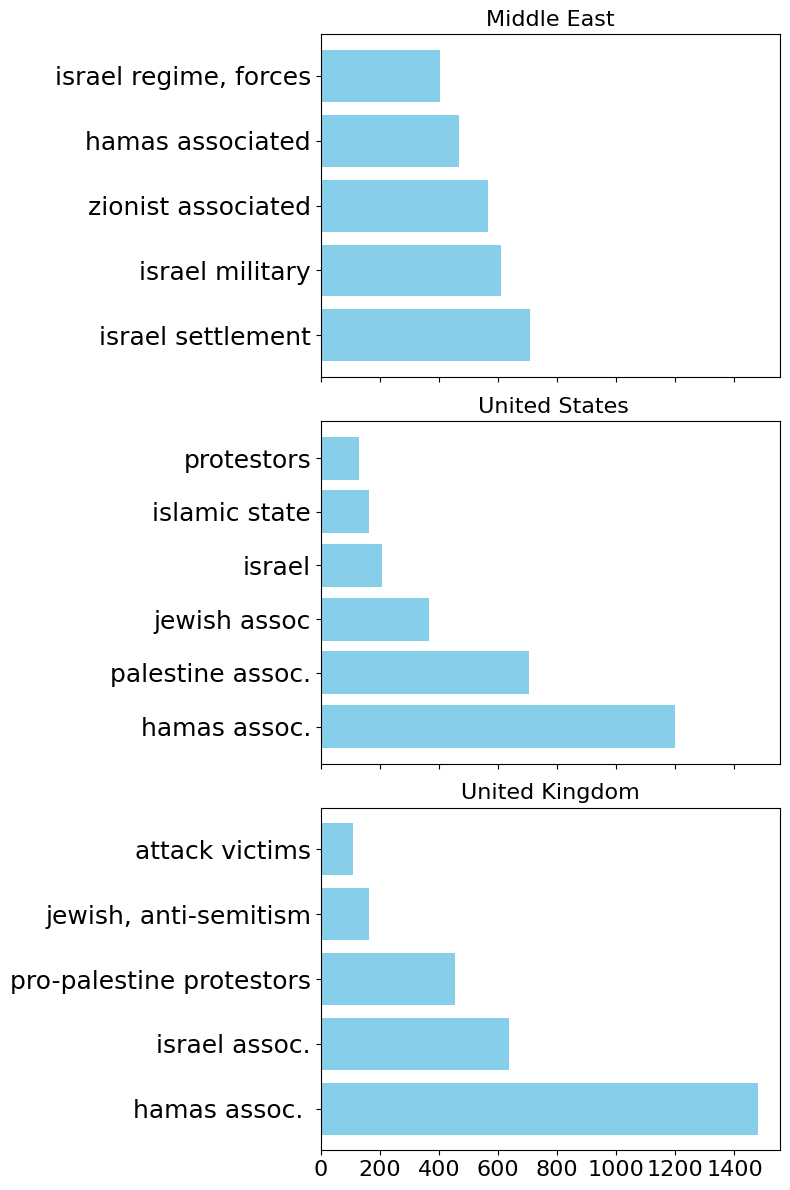}
   \caption{Most frequent targets of Demonizing and Dehumanizing Language across various regions. X-axis is the frequency of appearances. while the Y-axis are the target labels}
    \label{fig:demon}
\end{figure}

\paragraph{Language analysis}
In order to analyse instances of dehumanising and demonising language, we performed BERTopic clustering on the targets of such language in excerpts from the articles, obtained through the LLM, for the UK, US and Middle east. We collapse similar occurring instances into a single topic. \autoref{fig:demon} shows the the resulting counts of instances of dehumanizing and demonizing language used against specific clustered targets. We found that Hamas associated targets are consistently high across the UK and US. Israel-associated targets show a significant presence in the Middle East and United Kingdom. We repeated the same analysis for instances of victimising language and found that across ME, Palestinians are portrayed as victims in a significant proportion, as compared to the western counter parts, where Jewish community, and Israeli citizens and hostages, are portrayed as victims of the ongoing war.

\paragraph{Elite vs people focus}
In \autoref{tab:elite_v_people}, we demonstrate the differences in reporting across the regions. Overall frequencies for elite mentions are much higher compared to people mentions, demonstrating how war-centric the reporting is. This is consistent across all regions. Publishers from ME have more mentions in both categories compared to the other two. We can see ME publishers name Palestinians and Gazans much more often compared to the ones in the US or UK. Israel related terms are not explicitly mentioned in this context by ME publishers though, when they do make an appearance in the UK, US lists. 
Looking at the elite mentions, there is a higher focus on US related terms - president, Biden in the US and the UK, compared to the ME where Israel related terms are more frequent. Interestingly, Palestinian (from Palestinian authority) only shows up in the ME.

\begin{table*}[]
    \centering
    \small
\begin{tabular}{lll|lll}
\toprule
\multicolumn{3}{c}{People mentions} &
      \multicolumn{3}{c}{Elite mentions} \\
\midrule
US &                  UK &                  ME &                US &                UK &                  ME \\
\midrule
  (civilian, 24) &     (people, 29) &      (people, 56) & (president, 861) &  (minister, 518) &     (israeli, 881) \\
      (gaza, 20) &     (family, 12) & (palestinian, 54) &     (biden, 649) & (president, 465) &    (minister, 788) \\
    (people, 14) &     (impact, 12) &        (gaza, 52) &   (israeli, 487) &     (biden, 405) &   (president, 539) \\
    (impact, 12) &   (conflict, 11) &       (child, 37) &  (minister, 399) &   (israeli, 344) &     (foreign, 473) \\
   (hostage, 11) &   (civilian, 10) &    (civilian, 32) & (netanyahu, 354) &     (prime, 277) &           (us, 373) \\
    (family, 11) &    (martial, 10) &       (woman, 28) &     (house, 333) & (netanyahu, 276) &       (hamas, 351) \\
  (conflict, 11) &       (focus, 9) &      (impact, 21) &     (prime, 297) & (secretary, 272) &    (official, 325) \\
      (focus, 8) & (palestinian, 9) &      (killed, 21) &  (benjamin, 262) &       (joe, 255) &   (netanyahu, 315) \\
(palestinian, 8) &     (anthony, 9) &       (focus, 15) &    (israel, 252) &  (benjamin, 231) &       (prime, 284) \\
    (israeli, 7) &     (israeli, 8) &   (including, 14) &      (rep., 252) &    (leader, 211) & (palestinian, 282) \\
\bottomrule
\end{tabular}
    \caption{Most common words in instances of elite and people centric mentions across news articles per region}
    \label{tab:elite_v_people}
\end{table*}

\subsection{Semantic Framing}

\begin{figure}[h]
    \includegraphics[width=0.5\textwidth]{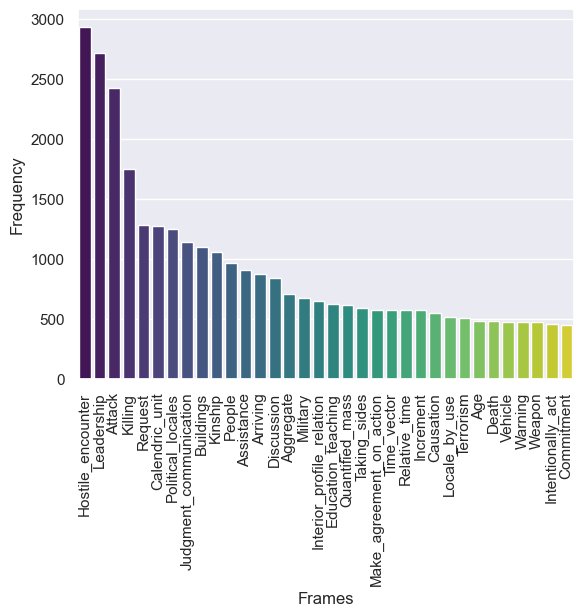}
    \caption{Relative frequency of Semantic Frames across the entire dataset}
    \label{fig:global-frames}
\end{figure}

The dominant semantic frames identified across all regions along with their relative frequencies are presented in Figure \ref{fig:global-frames}. From the figure, it can be inferred that there is a larger focus on semantic frames such as \textit{Hostile\_Encounter, attack, killing, political locales} pointing towards war coverage and a plausible presence between political locales, on either side. 

\begin{figure*}
\centering
   \includegraphics[width=0.9\textwidth]{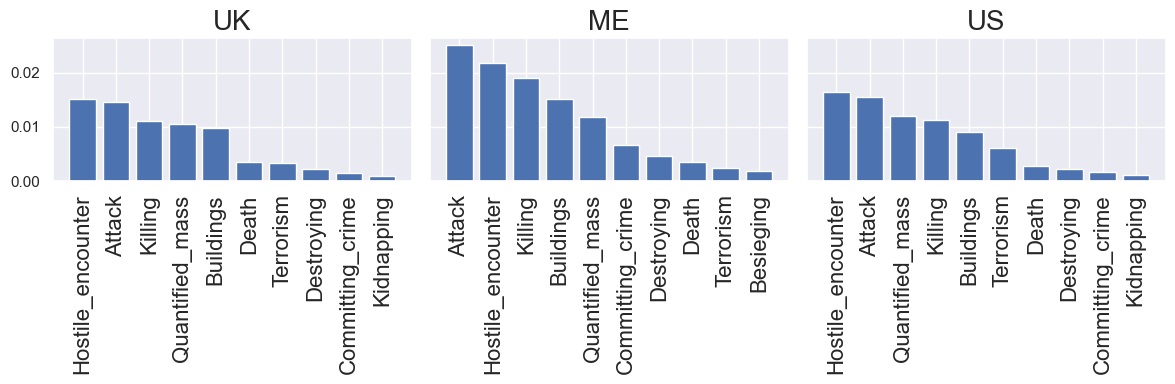}
   \caption{Visible Effects of war across regions}
   \label{fig:relative-frequencies-visible-effects}
\end{figure*}

\begin{figure*}
\centering
   \includegraphics[width=0.9\textwidth]{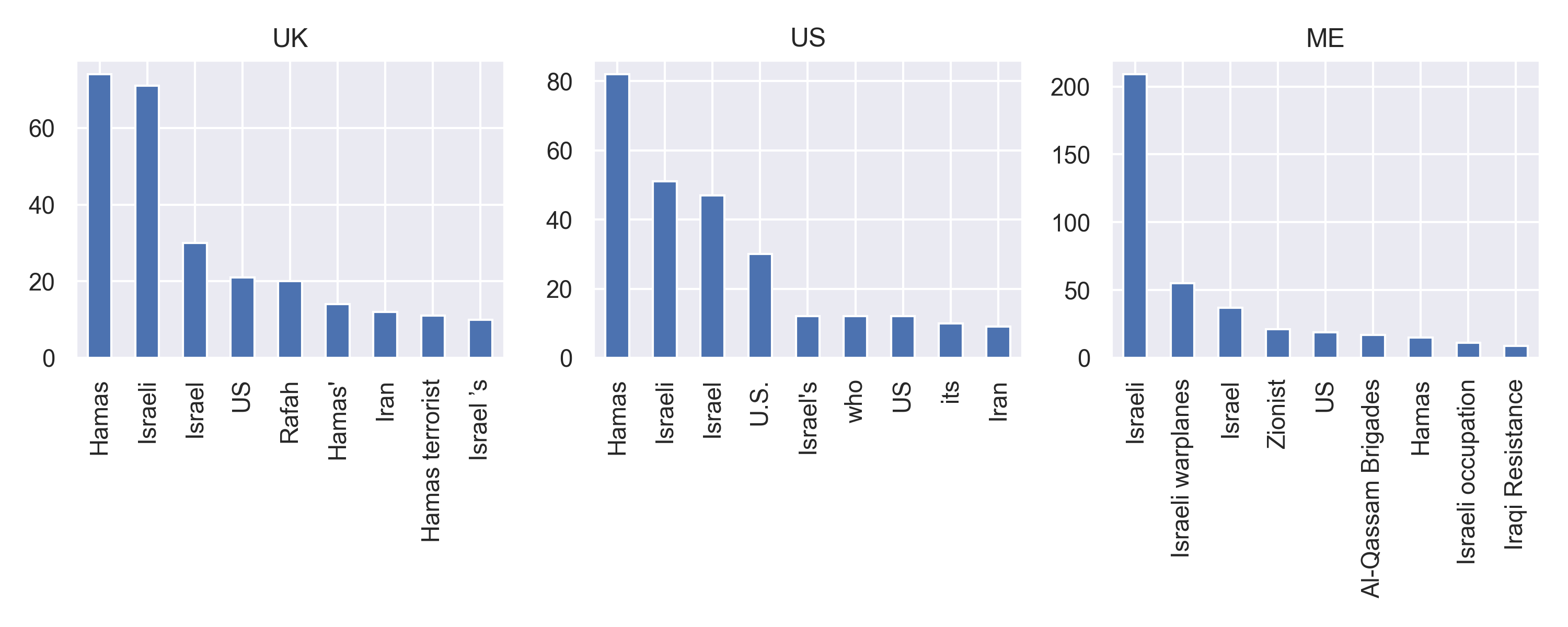}
   \caption{Relative frequencies of "Assailant" Frame Element for the "Killing" semantic frame across regions}
   \label{fig:assailants across regions}
\end{figure*}

\paragraph{ Visible and Invisible Effects of War}
\autoref{fig:relative-frequencies-visible-effects} shows that frames like \textit{"Attack", "Killing", "Hostile\_Encounter"} are dominant semantic frames present across all regions, denoting the visible effects of war. In case of invisible effects, we find that \textit{"Kinship", "Emotion\_directed", "Fear"} are commonly occurring frames. However, the proportion of visible effects of war is more than the invisible effects. We plot region-wise relative frequencies of visible effects of war in \autoref{fig:relative-frequencies-visible-effects}. We further investigate into the presence of this frame in individual regions. For example, we extract the frame elements for the frame \textit{"Attack"} . \textit{"Attack"} is described by frame elements such as \textit{'Assailant', 'Circumstances', 'Containing\_event', 'Depictive', 'Victim', 'Weapon'} etc.  On counting the occurrences of each of these frame elements associated with "Attack" in our filtered headlines, and find the elements \textit{"Assailant" and "Victim"} to be the most frequent amongst all. We find that across the US and UK, terms associated \textit{"Hamas"} fall more frequently in the assailant category as compared to \textit{"Israel"}, whereas in the Middle Eastern coverage, the opposite is found. The frame \textit{"Quantified\_mass"} to quantifies the number of people affected. We plot the co-occurrences of frames related to both visible and invisble effects of war across all regions in Figures \ref{fig:vis} and \ref{fig:invis} in the appendix. 

\section{Discussion and Conclusion}


In this work, we analyse reporting of the Israel-Palestine conflict in news media around the world. To that end, we collect a dataset of articles on the topic and analyse the media reporting through generic framing, issue-specific framing, and semantic framing, operationalising Galtung's framework for war and peace journalism~\citep{Galtung2013HighRL}. We use a combination of LLMs and a semantic framing model to identify both communicative and linguistic framing on the topic.

Through our analysis, we capture differences in the use of language in war reporting across various media houses and regions. We find that adversarial framing of groups, the use of demonizing, dehumanizing language is much more common and there is a higher focus on visible effects of war. Further, media sources from different regions portray different entities as the assailants and victims of the crimes being committed. Publishers in the Middle East tend to have a much higher focus on health, people's security, and regulation while publishers in the US, UK have a higher focus on public opinion on the topic and how politics is involved.

Such higher focus on conflict-oriented framing rather than solution-oriented and regional differences in depicting who the assailants and victims are can worsen the conflict and create more divide in society. There have been several studies in the past linking media framing to effects in the real world. Excessive use of aggressive language in media reporting can have adverse psychological effects on readers such as incitement of hate~\citep{kovalchuk2022psycholinguistic}. Studies have also shown subtle linguistic differences in news reporting can influence whether people interpret violent acts as patriotism or terrorism~\citep{dunn2005war}. It has been found that long term exposure to such derogatory language can lead to political radicalization among groups, reduces people's ability to recognize its offensive character and shape the way they construct reality~\citep{bilewicz2020hate, doi:10.1177/0261927X20967394}.
Beyond journalism, media bias in this form also has an impact associated to LLMs. News articles are a primary source of training of language models, which are known to amplify the biases present in their training data~\citep{feng-etal-2023-pretraining, wright-etal-2024-llm}. These, in turn, can then have an impact on people having interactions with these models~\citep{10.1145/3544548.3581196} and shape their opinion, which could cause inadvertent harm to individuals and society at large.

Galtung in his work throws light upon the fact that focus on the violence is equivalent to taking "low road" in journalism, specifically in the context of war reporting. This is because it tends to encourage more violence, and creates an \textit{us vs them} mindset, focused on a zero-sum or victory orientation. During times of war, taking such a "low road" in media, should be refrained from and actively criticized, to prevent further escalation and polarization. It is therefore essential to change the discourse within which conflict is covered in media. During a conflict, there is a need to depolarize, de-escalate by emphasizing invisible effects of war, the emotional turmoil suffered, voicing diverse perspectives, and ultimately humanizing all sides in conflict. This in turn will result in charting a course for solution and peace oriented journalism.
In the words of \citet{Galtung2013HighRL},\textit{"Good reporting on conflict is not a compromise, a little from the left hand column, a little from the right but favours peace journalism and opposes war journalism.  Do not corrupt the media by giving the task to them, having them take it on voluntarily, or forcing them into that kind of journalism..."}
By operationalising Galtungs framework using NLP, we hope that our analysis can contribute towards highlighting the need to step away from war-oriented journalism, paving the way for a solution seeking discourse.
 


\section{Limitations}
Our study is not without limitations. One of the limitations, is that we used only english media sources in our study. In order to get a fine-grained understanding of discourse surrounding the conflict, it is essential to also consider news in regional languages. 
Furthermore, we have used large language models and transformer based seq-to-seq models to identify issue specific frames. These models are known to pick up biases in training data, thereby biasing the results. Another limitation is the the size of the data for our analysis, which is limited. In practice, it would have been nicer to have them across even more diverse media sources and larger data set size.

\bibliography{anthology,custom}
\bibliographystyle{acl_natbib}

\appendix

\section{MFC evaluation}

We report the per label metrics of the generic frame classifer on the MFC dataset in \autoref{tab:text-frame-perf-label}. 

\begin{table}[]
\begin{tabular}{llll}
\toprule
Label & Precision & Recall & F1-score \\
\midrule
cap\&res      & 0.39   & 0.34     & 0.36 \\
crime         & 0.50   & 0.87     & 0.63 \\
culture       & 0.38   & 0.37     & 0.37 \\
economic      & 0.43   & 0.69     & 0.53 \\
fairness      & 0.17   & 0.74     & 0.28 \\
health        & 0.48   & 0.48     & 0.48 \\
legality      & 0.53   & 0.87     & 0.66 \\
morality      & 0.30   & 0.63     & 0.41 \\
policy        & 0.40   & 0.73     & 0.51 \\
political     & 0.68   & 0.53     & 0.60 \\
public\_op    & 0.32   & 0.55     & 0.40 \\
quality\_life & 0.28   & 0.36     & 0.31 \\
regulation    & 0.26   & 0.48     & 0.34 \\
security      & 0.30   & 0.45     & 0.36 \\
\midrule
micro avg     & 0.42   & 0.62     & 0.50 \\
macro avg     & 0.39   & 0.58     & 0.45 \\
weighted avg  & 0.45   & 0.62     & 0.51 \\
samples avg   & 0.46   & 0.63     & 0.51 \\
\bottomrule
\end{tabular}
\caption{Metrics per frame label (multi-label) for the text frame classifier on the MFC dataset from \citet{arora2025multimodalframinganalysisnews}}
\label{tab:text-frame-perf-label}
\end{table}

\section{Additional Figures}
\begin{figure}[ht]
    \centering    \includegraphics[width=\columnwidth]{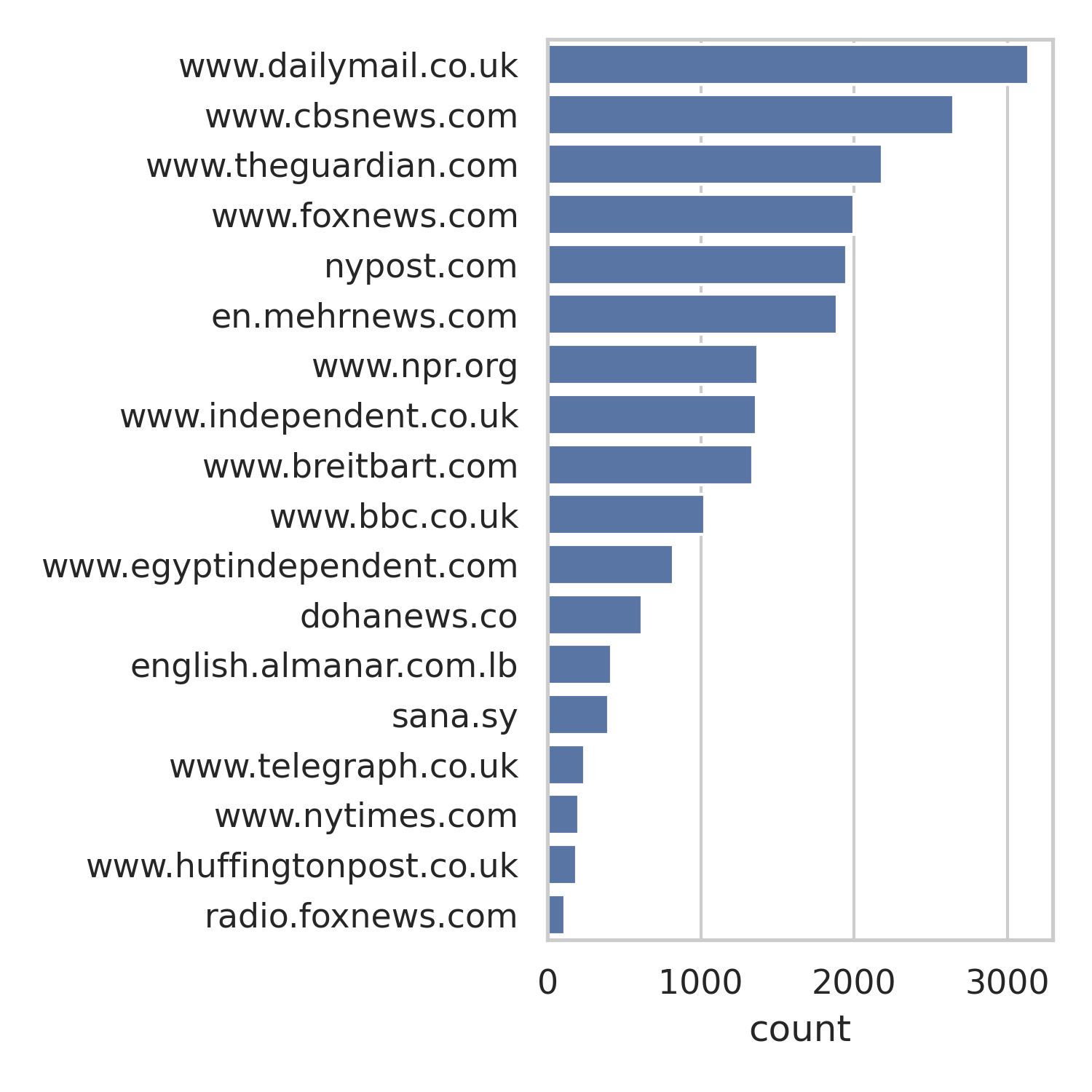}
    \caption{Number of articles for each domain in the filtered dataset}
    \label{fig:freq_sources}
\end{figure}

\begin{figure*}[ht!]
    \centering    \includegraphics[width=\textwidth]{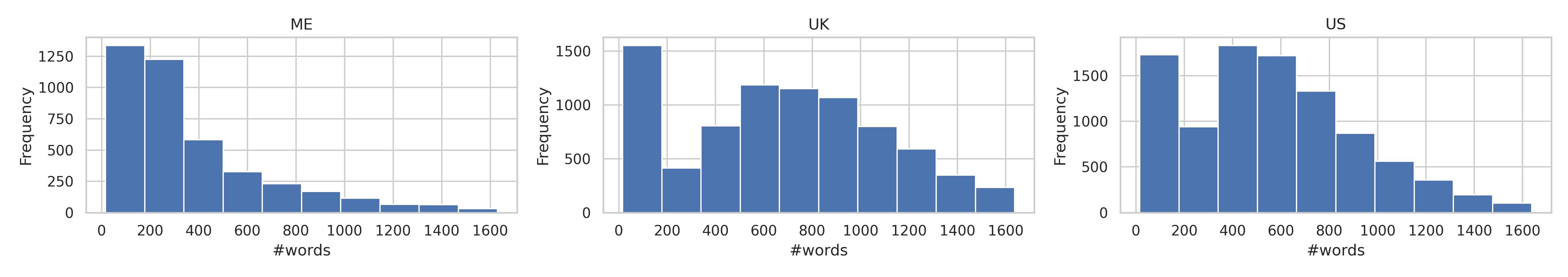}
    \caption{Distribution of length of articles from each region}
    \label{fig:data-stat-len}
\end{figure*}

\begin{figure*}[ht!]
    \centering    \includegraphics[width=\textwidth]{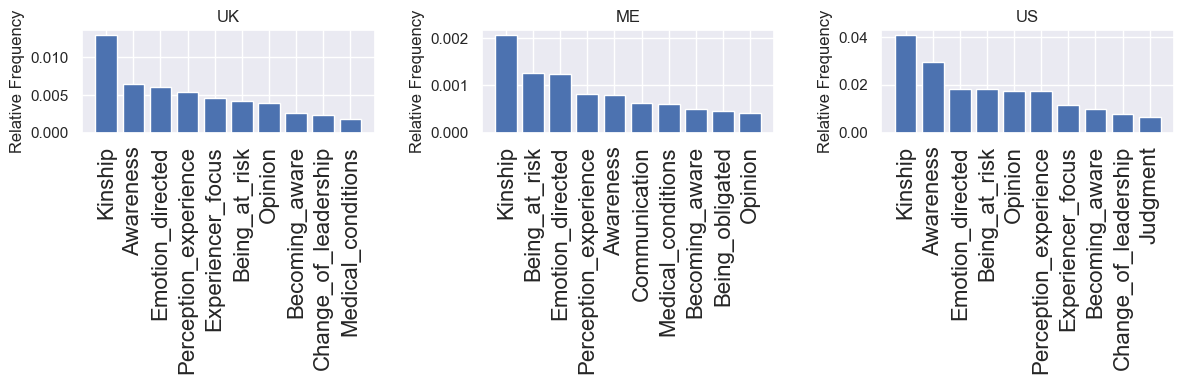}
    \caption{Semantic frames showing invisible effects of war across UK, US, ME}
    \label{fig:invisible-effects-of-war}
\end{figure*}

\begin{figure*}[ht!]
    \centering    \includegraphics[width=\textwidth]{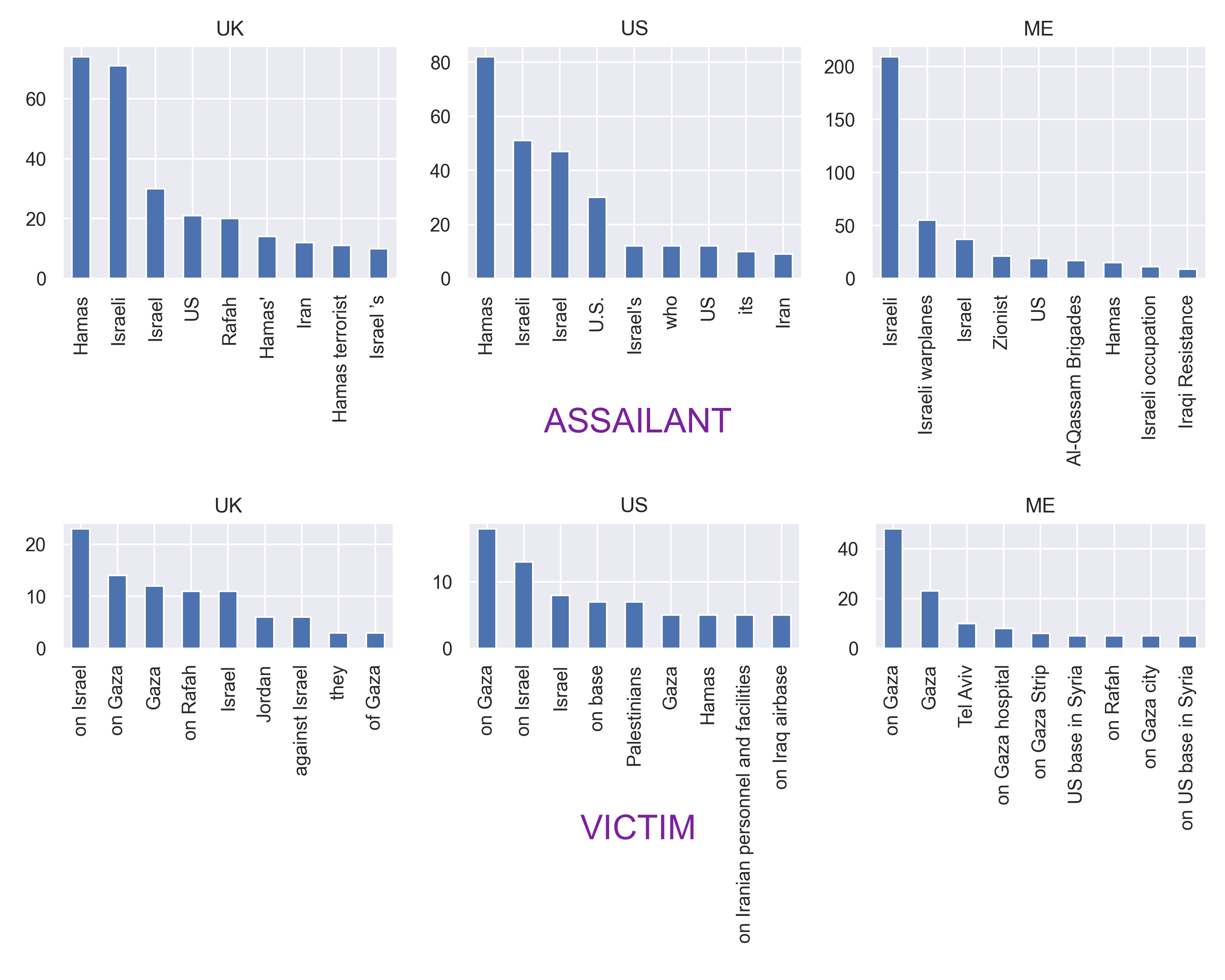}
    \caption{Assailants and Victims in headlines across UK, US, ME}
    \label{fig:invisible-effects-of-war}
\end{figure*}

\begin{figure*}[h]
    \centering
    \begin{subfigure}[b]{0.3\textwidth}
        \centering
        \includegraphics[width=\linewidth]{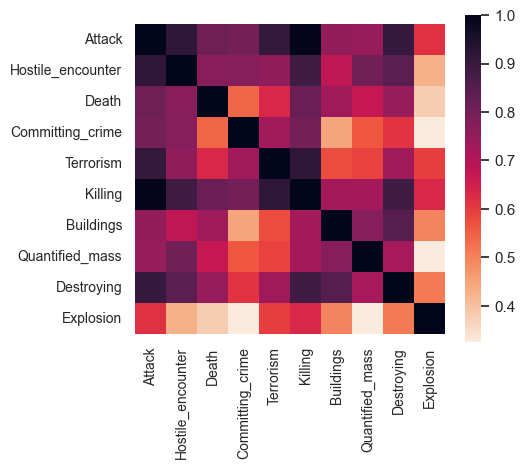}
        \caption{Co-Occurring semantic frames in the UK}
        \label{fig:uk_coccur_vis}
    \end{subfigure}
    \hfill 
    \begin{subfigure}[b]{0.3\textwidth}
        \centering
        \includegraphics[width=\linewidth]{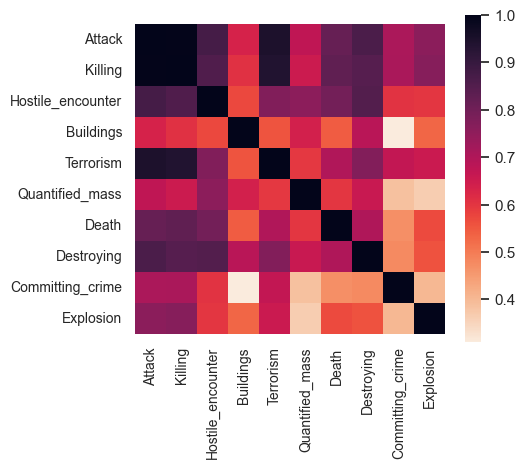}
        \caption{Co-Occurring semantic frames in the US}
        \label{fig:us_coccur_vis}
    \end{subfigure}
    \hfill 
    \begin{subfigure}[b]{0.3\textwidth}
        \centering
        \includegraphics[width=\linewidth]{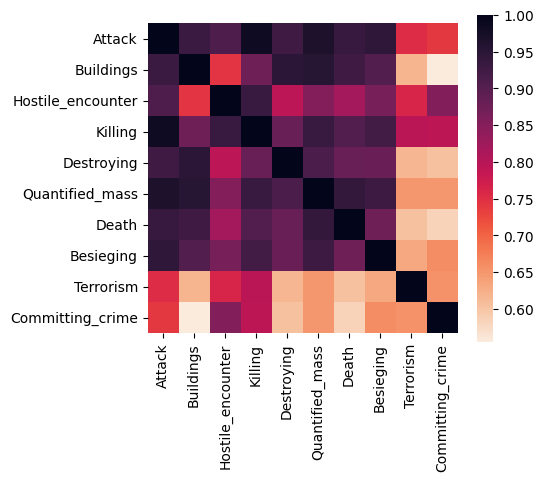}
        \caption{Co-Occurring semantic frames in the Middle East}
        \label{fig:me_coccur_vis}
    \end{subfigure}
    \caption{Visible Effects of War}
    \label{fig:vis}
\end{figure*}

\begin{figure*}[h]
    \centering
    \begin{subfigure}[b]{0.32\textwidth}
        \centering
        \includegraphics[width=\linewidth]{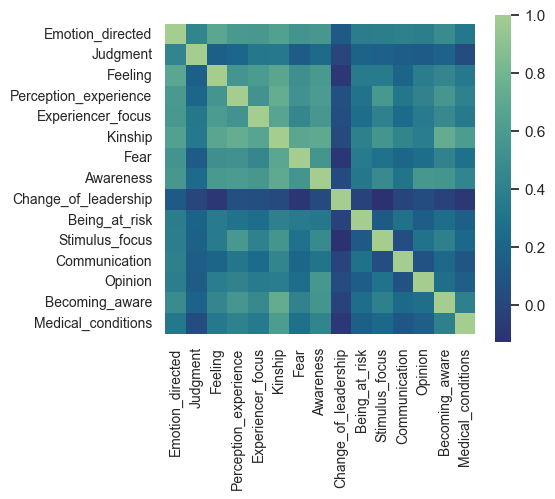}
        \caption{Co-Occurring semantic frames in the UK}
        \label{fig:uk_invis}
    \end{subfigure}
    \hfill 
    \begin{subfigure}[b]{0.32\textwidth}
        \centering
        \includegraphics[width=\linewidth]{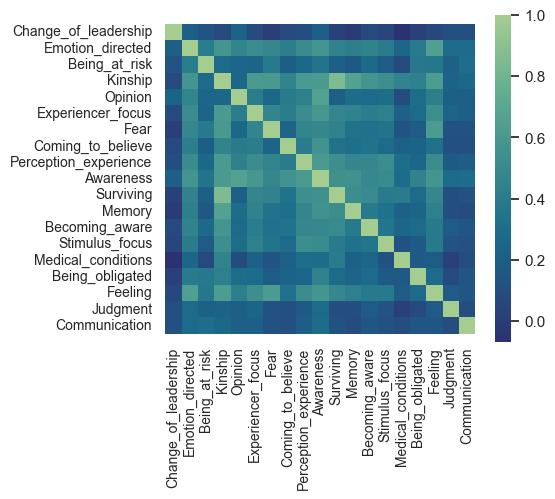}
        \caption{Co-Occurring semantic frames in the US}
        \label{fig:us_invis}
    \end{subfigure}
    \hfill 
    \begin{subfigure}[b]{0.32\textwidth}
        \centering
        \includegraphics[width=\linewidth]{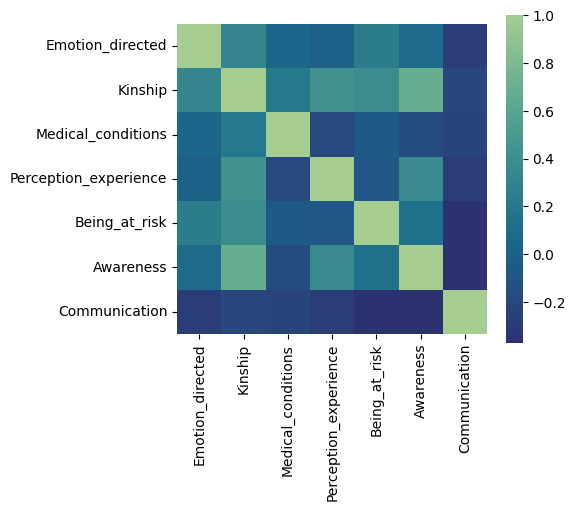}
        \caption{Co-Occurring semantic frames in the Middle East}
        \label{fig:me_invis}
    \end{subfigure}
    \caption{Invisible Effects of War}
    \label{fig:invis}
\end{figure*}

\begin{figure*}[h]
    \centering
    \begin{subfigure}[b]{\textwidth}
    \centering
        \includegraphics[width=\linewidth]{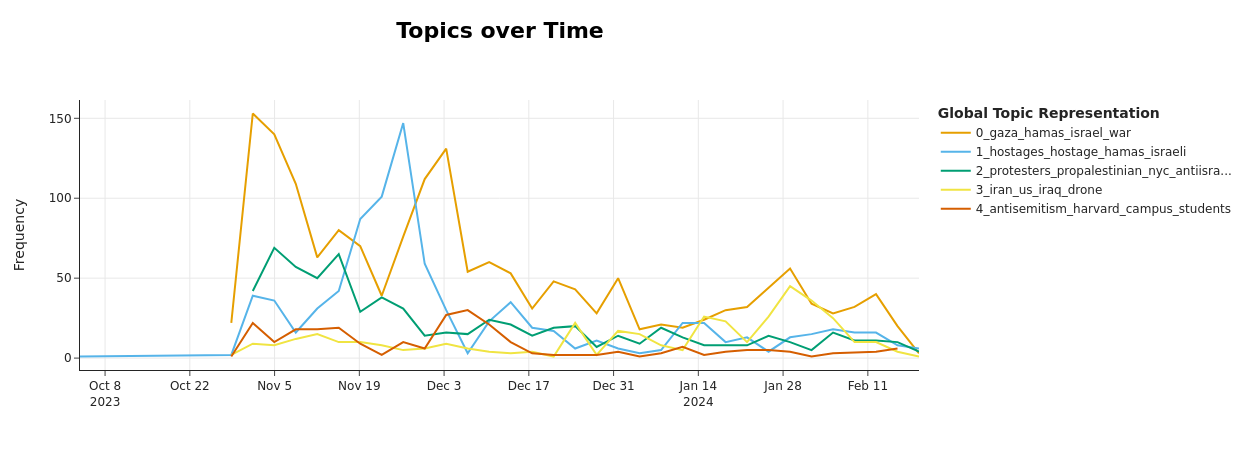}
        \label{fig:US-topics}
    \end{subfigure}
    \vfill 
    \begin{subfigure}[b]{\textwidth}
    \centering
        \includegraphics[width=\textwidth]{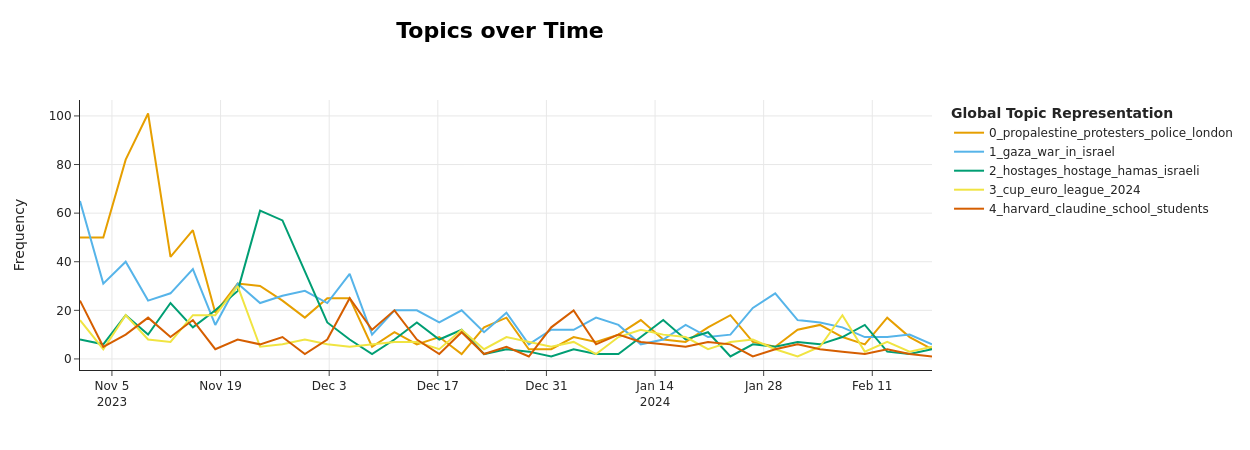}
        \label{fig:UK-topics}
    \end{subfigure}
    \vfill 
    \begin{subfigure}[b]{\textwidth}
    \centering
        \includegraphics[width=\textwidth]{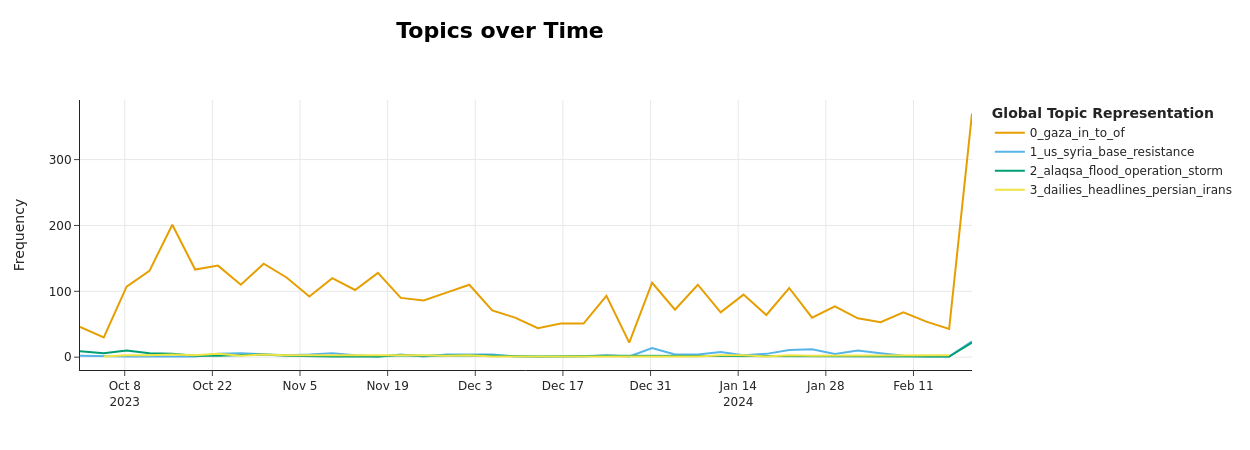}
        \label{fig:ME-topics}
    \end{subfigure}
    \caption{Topics over time in reporting from UK, US, and Middle Eastern sources}
    \label{fig:invis}
\end{figure*}

\begin{table*}[h]
\centering
\small
\begin{tabular}{>{\raggedright}p{2cm}p{5cm}|p{2.5cm}p{5.5cm}}
\toprule
\multicolumn{2}{c}{\textbf{Visible Effects}} & \multicolumn{2}{c}{\textbf{Invisible Effects}}\\
\midrule
\textbf{FrameNet Frame} & \textbf{Definition}  & \textbf{FrameNet Frame} & \textbf{Definition}\\
\midrule
Hostile\_encounter & Aggressive interactions between/w two or more parties, physical violence or military action. & People and People\_by\_Age & Human aspect of conflicts focusing on how different demographics (children, adults, the elderly) are  affected \\
\addlinespace
Attack & Assailant or an aggressor attacking a target physically, metaphorically, involving conflict. & Emotion\_Directed and Fear & Emotional climate of a society during and after conflicts. Fear, in particular, can indicate areas of instability and the psychological impacts of war, which can persist long after physical conflicts have ended.  \\
\addlinespace
Killing & Killer causing the death of a victim & Kinship & familial structures and relationships are affected by war - Displacement, loss, and separation can alter kinship dynamics significantly, affecting social cohesion and community support systems. \\
\addlinespace
Quantified\_mass &  A number, quantifiable unit & Medical\_Conditions & health consequences of war, not only due to immediate injuries but also due to long-term psychological and physiological stress, malnutrition, and the breakdown of healthcare systems \\
\addlinespace
Military\_operation &  The military force (nation, group, individual) etc directed towards achieving some military goals. & Assistance and Awareness &  response to war - affected populations and externally by international actors  \\
\addlinespace
Building &  Can aid in identifying visible damage to economic property. & Being\_at\_Risk and Casualties & Human cost of conflict\\
\addlinespace
Terrorism & A violent and harmful act in order to terrorize a population/ govt.  \\
\addlinespace
Death & Death of a protagonist, along with its cause.  \\
\addlinespace
Destroying & A Destroyer or Cause \\
\addlinespace
Committing\_crime & A Perpetrator committing a crime not permitted by law and society \\
\addlinespace
BuildFiring &  Can aid in identifying visible damage to economic property. \\
\bottomrule
\end{tabular}
\caption{Visible and Invisible Effects of War (non-exhaustive): Frames of Interest and their definitions}
\label{tab:frames}
\end{table*}

\begin{figure*}
\begin{lstlisting}[basicstyle=\tiny, style=mystyle]
SYS_PROMPT ="""
You are a helpful AI assistant.
"""

prompt_template = """
Given an article as input your task is to analyse it along the framework for Galtung's War and Peace journalism framework.

You have to assess the framing of the article, and come up with salient indicators supporting war or peace journalism frames, according to the framework.
Indicators include attribution of blame, partisan framing, the reporting being elite oriented vs people-oriented, or the language of the article being victimizing, demonizing or dehumanizing a certain group.
List the indicators found and provide exact phrasing of each indicator from the article. Identify the targets of the indicator, and give an associated reasoning.

Structure your output in a json format, with each indicator as key and the corresponding wording of that indicator, targets, and associated reasoning in the article text as the values in a list format.
```json

{{
  "war_journalism_indicators": {{
    "adversarial_frame": {{
      "use_of_adversarial_language": [(<List of instances from the article>, <target>, <reasoning>)],
      "attribution_of_blame": [(<List of instances from the article>, <target>,  <reasoning>)],
    }},
    "focus_on_elites": [<List of instances from the article>],
    "attribution_of_blame": [(List of instances from the article, <target>, <reasoning>)],
    "labelling_of_people": [(List of instances from the article, <target>, <reasoning>)],
    "language": {{
      "demonizing_language": [(<List of instances from the article>, <target>, <reasoning>)],
      "dehumanizing_language": [(<List of instances from the article>, <target>, <reasoning>)],
      "victimizing_language": [(<List of instances from the article>, <target>, <reasoning>)],
      "passive_language": [(<List of instances from the article>, <target>, <reasoning>)]
    }},
    "partisan_framing": [(<List of instances from the article>, <target>, <reasoning>)],
    "focus_on_visible_effects_of_war":[<List of instances from the article>],
    "nationalistic_frame": {{
      "emphasis_on_national_interests": [(<List of instances from the article>, <target>, <reasoning>)],
      "portrayal_of_national_strength": [(<List of instances from the article>, <target>, <reasoning>)]
    }},
    "military_solution": [<List of instances from the article>],
  }},
  "peace_journalism_indicator": {{
    "peace_frame": {{
      "focus_on_consequences_of_conflict": [(<List of instances from the article>, <target>, <reasoning>)],
      "inclusion_of_peace_proposals": [(<List of instances from the article>, <target>, <reasoning>)],
      "representation_of_multiple_perspectives": [(<List of instances from the article>, <target>, <reasoning>)]
    }},
    "focus_on_invisible_effects_of_war":[(<List of instances from the article>, <target>)],
    "peace_orientation": [(<List of instances from the article>, <target>, <reasoning>)],
    "people_orientation":[(<List of instances from the article>, <target>, <reasoning>)],
    "victim_orientation": [(<List of instances from the article>, <target>, <reasoning>)]
  }}
}}
Article: {article}
\end{lstlisting}
\end{figure*}

\begin{figure*}
\begin{lstlisting}[basicstyle=\tiny, style=mystyle]

SYS_PROMPT = f"You are an intelligent and logical journalism scholar conducting analysis of news articles. Your task is to read the article and answer the following question about the article. Only output the json and no other text.\n"

FRAMES = """
A list of frame names and their descriptions used in news is:
Economic - costs, benefits, or other financial implications,
Capacity and resources - availability of physical, human, or financial resources, and capacity of current systems, 
Morality - religious or ethical implications,
Fairness and equality - balance or distribution of rights, responsibilities, and resources,
Legality, constitutionality and jurispudence - rights, freedoms, and authority of individuals, corporations, and government,
Policy prescription and evaluation - discussion of specific policies aimed at addressing problems,
Crime and punishment - effectiveness and implications of laws and their enforcement,
Security and defense - threats to welfare of the individual, community, or nation,
Health and safety - health care, sanitation, public safety,
Quality of life - threats and opportunities for the individual's wealth, happiness, and well-being,
Cultural identity - traditions, customs, or values of a social group in relation to a policy issue,
Public Opinion - attitudes and opinions of the general public, including polling and demographics,
Political - considerations related to politics and politicians, including lobbying, elections, and attempts to sway voters,
External regulation and reputation - international reputation or foreign policy of the U.S,
None - none of the above or any frame not covered by the above categories."""
    
GENERIC_FRAMING_MULTIPLE_PROMPT = """
Given the list of news frames, and the news article.
Your task is to carefully analyse the article and choose the appropriate frames used in the article from the above list.
Output your answer in a json format with the format:
{"frames-list": "[<All frame names that apply from list provided above>], "reason": "<reasoning for the frames chosen>"}.
Only choose the frames from the provided list of frames. If none of the frames apply, output "None" as the answer.
"""
\end{lstlisting}
\end{figure*}

\end{document}